\documentclass[conference,letterpaper]{IEEEtran}

\usepackage{graphics} 
\usepackage{epsfig} 

\usepackage{mathptmx} 
\usepackage{times} 
\usepackage{amsmath} 
\usepackage{amssymb}  
\usepackage{bm}
\usepackage{euscript}

\usepackage[tight,footnotesize]{subfigure}
\usepackage{graphicx}
\usepackage{mdwmath}
\usepackage{bbm}
\usepackage[table]{xcolor}
\usepackage{multirow}
\usepackage{tabularx}
\usepackage{booktabs}

\usepackage{algorithm}
\usepackage{algorithmic}
\usepackage{url}

\usepackage{comment}
\usepackage{makecell}

\usepackage{cite}
\newcommand{\ie}{i.e.}

\newcommand{\equ}[1]{(\ref{eq:#1})}

\begin{document}

\title{High-resolution LIDAR-based Depth Mapping using Bilateral Filter}

\author{\IEEEauthorblockN{Cristiano Premebida\IEEEauthorrefmark{1}\IEEEauthorrefmark{2},
Luis Garrote\IEEEauthorrefmark{1}\IEEEauthorrefmark{2},
Alireza Asvadi\IEEEauthorrefmark{1}\IEEEauthorrefmark{2},
A. Pedro Ribeiro\IEEEauthorrefmark{1}, and
Urbano Nunes\IEEEauthorrefmark{1}\IEEEauthorrefmark{2}}
\IEEEauthorblockA{\IEEEauthorrefmark{1}Department of Electrical and Computer Engineering (DEEC)}
\IEEEauthorblockA{\IEEEauthorrefmark{2}Institute of Systems and Robotics (ISR)\\
University of Coimbra, Portugal.\\
Emails: {\tt\small\{cpremebida,garrote,asvadi,urbano\}@isr.uc.pt}}
}

\maketitle

\begin{abstract}
High resolution depth-maps, obtained by upsampling sparse range data from a 3D-LIDAR, find applications in many fields ranging from sensory perception to semantic segmentation and object detection. Upsampling is often based on combining data from a monocular camera to compensate the low-resolution of a LIDAR. This paper, on the other hand, introduces a novel framework to obtain dense depth-map solely from a single LIDAR point cloud; which is a research direction that has been barely explored. The formulation behind the proposed depth-mapping process relies on local spatial interpolation, using sliding-window (mask) technique, and on the Bilateral Filter (BF) where the variable of interest, the distance from the sensor, is considered in the interpolation problem. In particular, the BF is conveniently modified to perform depth-map upsampling such that the edges (foreground-background discontinuities) are better preserved by means of a proposed method which influences the range-based weighting term. Other 
methods for spatial upsampling are discussed, evaluated and compared in terms of different error measures. This paper also researches the role of the mask's size in the performance of the implemented methods. Quantitative and qualitative results from experiments on the KITTI Database, using LIDAR point clouds only, show very satisfactory performance of the approach introduced in this work.
\end{abstract}

\IEEEpeerreviewmaketitle
\section{Introduction}
\label{sec:introduction}

\begin{figure}[!t]
\centering
\includegraphics[width= 89.0 mm]{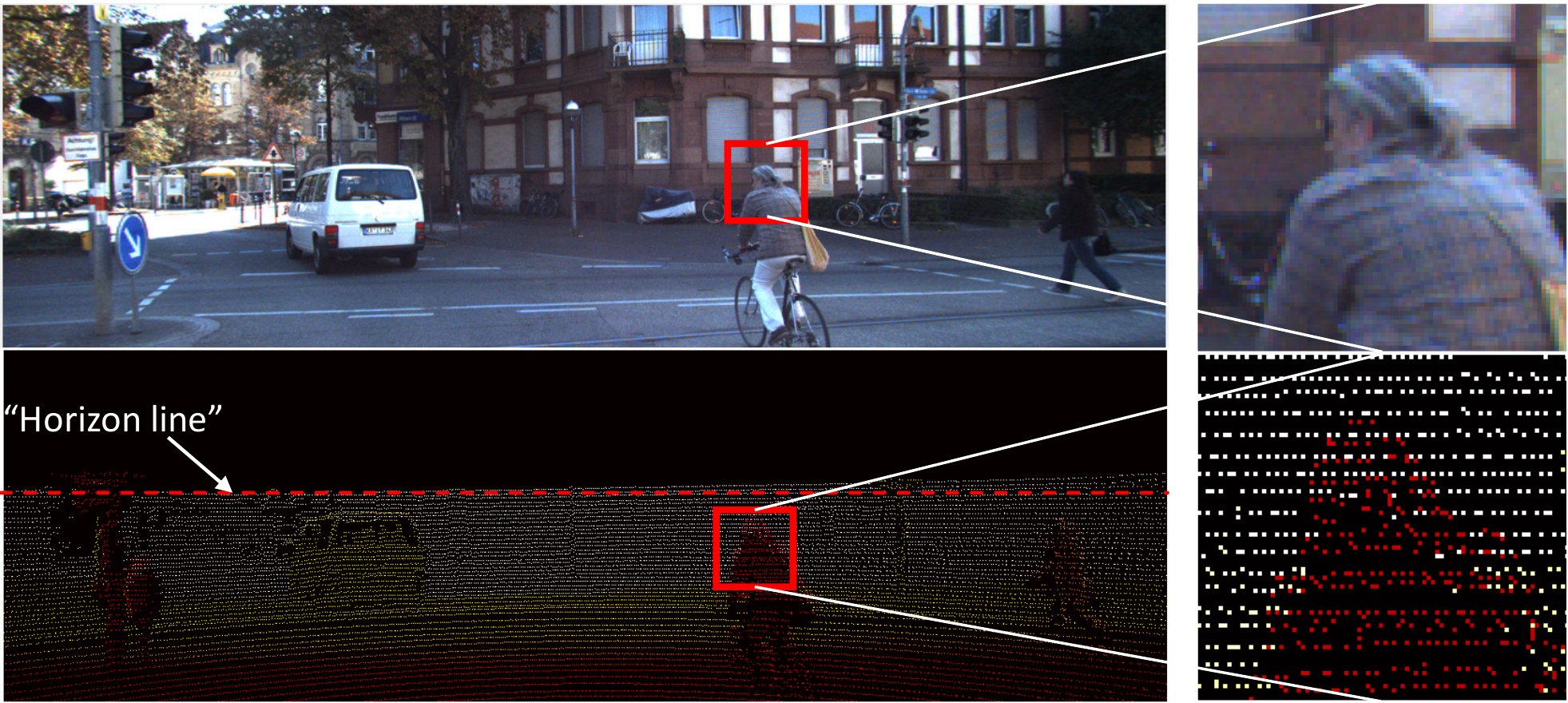}
\caption{Example of an urban scene (image on the top) where the LIDAR point-cloud is shown in pixel coordinates (bottom). The sparseness of the LIDAR is emphasized in the zoom view at the bottom-right. This frame is part of the KITTI-dataset, where the LIDAR (a Velodyne HDL-64E) is mounted on the roof of a vehicle and the LIDAR covers a field-of-view (FOV) limited by a vertical angle - approximated by a line (red dashed). This line will be hereafter called horizon line.}
\label{fig:1}
\end{figure}

This paper deals with the problem of obtaining a depth map, in pixel coordinates, from a single 3D point-cloud generated by a multi-channel LIDAR mounted on-board an instrumented vehicle. Assuming the LIDAR is calibrated wrt a monocular camera, the transformation of the point-cloud in $\mathbb{R}^3$ to the image plane generates a sparse distribution of points, as shown in Fig. \ref{fig:1} (zoom at the bottom-right), where the majority (more than 90\%) of pixels are unsampled. Some difficulties arise in obtaining a consistent and dense (high resolution) depth map, where consistency has to do with obtaining a good estimation of depth values in edges and smooth regions of the depth-map. On the other hand, density is related to the spatial resolution of the map where the problem is caused by sparse and incomplete data. Dense depth maps are applicable in several fields, such as artificial perception, sensor fusion, scene reconstruction and semantic segmentation. Upsampling sparse 3D point-clouds to obtain a high 
resolution depth image can be explored in the context of artificial sensory-perception for intelligent/autonomous vehicles and ADAS applications \cite{Broggi}, such as: road and obstacle detection \cite{Wolf}, curb detection \cite{Stiller}, vehicle detection \cite{Zhang}.

The majority of the works in this research area use, in conjunction with range data, information from monocular camera as in \cite{Dolson,Triebel,YHe,Miksik}. This work, instead, differs from existing solutions because the proposed approach uses only data from a LIDAR, and therefore monocular camera is considered just for calibration and visualization purposes. More specifically, a Bilateral Filter (BF) based framework is described which generates, from single 3D point-cloud delivered by a Velodyne HDL-64, a high resolution depth-map. Upsampling 3D point-clouds to produce a dense depth map based solely on data provided by a LIDAR \ie, color/texture information from camera is not been used, is a research area with very few scientific literature. The recent work of Miksik \textit{et al.} \cite{Miksik} shares some common points with the approach described in this paper but, data from a stereo-image system is used in their algorithm. Conversely, methods that combine data from LIDAR and camera \ie, the depth 
upsampling strategy is enhanced by color and texture information, are much more common as evidenced by a number of interesting scientific works, such as \cite{Dolson}, \cite{Triebel}, \cite{AGonzalez}. Besides the problem of the low number of sampled points provided by a single point-cloud, which corresponds to less than 10\% of the region within the FOV (see Fig. \ref{fig:1}), particular attention has to be given to the ambiguity problem of foreground vs background, that occurs in the areas of discontinuities (edges) between objects (such as a pedestrian or a vehicle) and far away objects (background). This is an aspect that will be addressed in detail in Sect \ref{sec:Filtering} and is one of the main reasons to use edge-preserving methods like the BF. This work utilizes the structure of edge-preserving filters, namely the BF, which means that the sampled points in the local window are weighted by combining a neighborhood distance function and a range based function. The first weighting function does not 
contribute significantly to the performance of the system, while the second function plays a key role on the final results. These aspects will be further discussed and demonstrated. 

Edge-preserving filters are very popular in the computer vision community, and they are used in many applications such as noise reduction, stereo matching, image deconvolution, image upsampling. Examples of this type of filter are the Anisotropic Diffusion filter, the Bilateral filter, and the recent Guided filter. The Anisotropic Diffusion filter was proposed by Perona and Malik \cite{Perona} to be a method to preserve region boundaries (edges) of objects in images. One of the criteria enunciated in \cite{Perona} is that \textit{the region boundaries should be sharp and coincide with the semantically meaningful boundaries}. Their idea was to prioritize smoothing within a local region over to smoothing across boundaries. On the other hand, the BF, named in \cite{Tomasi}, gained much popularity and demonstrated to be a very useful 
and efficient method in many applications. Examples of works which employed the BF to obtain dense depth-maps from LIDAR data are \cite{Dolson}, \cite{FusionIROS14}. Recently, the Guided filter \cite{GuidedFilter} was proposed as an alternative to the BF in the sense that it is also an edge-preserving smoothing filter \cite{GuidedFilter}. However, so far the Guided filter has been used only in the domain of computer vision applications and, therefore, work on LIDAR data upsampling seems to be an open problem to be addressed.

In terms of contributions, this paper proposes an upsampling framework for high-resolution depth mapping based on the BF, using an original approach to process the range values, within a local mask, where the discontinuities (edges) are meaningfully preserved. The proposed approach is particularly suitable for LIDAR only depth maps. Moreover, this work provides a thorough experimental validation of the framework, considering qualitative and quantitative criteria, and reports experimental results and comparative evaluation with several other techniques.

The remainder of this paper is organized as follows, in the next section the problem formulation is presented, Sect. \ref{sec:Filtering} is devoted to edge-preserving filtering, while Sect. \ref{sec:modified_bilateral} focuses on the proposed solution to the discontinuity problem using BF. Sect. \ref{sec:eval} describes the evaluation methodology and datasets; experiments and results are detailed in Sect. \ref{sec:Experiments}, and finally Sect. \ref{sec:conclusions} concludes this paper.

\section{Depth-map upsampling formulation}
\label{sec:sect1}

Given a 3D point-cloud $PC^{t} \subset \mathbb{R}^3$ produced by a LIDAR at a given time-stamp $t$ (from now on $t$-index will be omitted to simplify the notations), and assuming the LIDAR is calibrated wrt a monocular camera with image-plane denoted by $\Pi_{2}$, the formulation starts by considering the set of points $P \in PC$ that lie within $\Pi_{2}$ \ie, $P$ is obtained after the transformation from  $\mathbb{R}^3$ to the camera coordinate system and then to the image-plane using the calibration matrix. Given the set of points $P \in \Pi_{2}$, with $P$ having non-integer pixel coordinates and being sparse, the goal here is to obtain a high resolution depth-map $DM$ (high density of points) limited by the size of $\Pi_{2}$ ($mu \times mv$) and by the horizon line. In terms of locations, $DM$ is restricted to positive-integer values \ie, pixel locations, in the range $[0,1,\cdots, mv ]$ and $[0,1,\cdots,mu ]$ respectively for the horizontal and vertical positions.

\begin{figure}[!t]
\centering
\includegraphics[width= 91.0 mm]{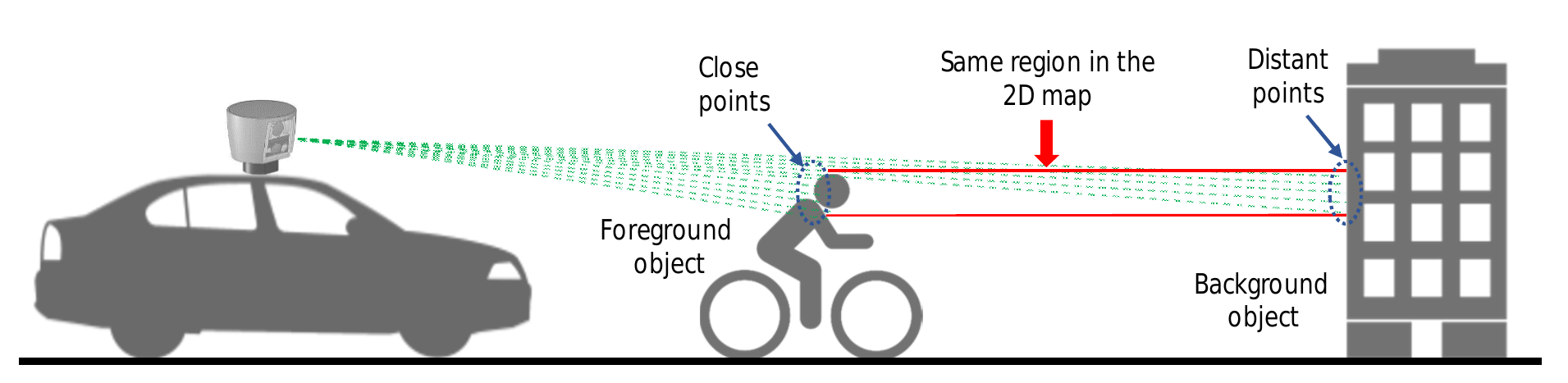}
\caption{Illustration of the superposition between close (foreground) and far (background) objects, whose measurement points will be located in the same local (vicinity) region of the depth-map.}
\label{fig:foreground_back}
\end{figure}

The elements of $P$ are the points $\{ \mathbf{p_1}, \cdots, \mathbf{p_n} \}$, where each point $\mathbf{p_i} = (u,v,r)_i$ is represented by the position in pixel coordinates $(u,v)_i$ and by the range value $r_i$ as measured by the LIDAR. However, since $u$ and $v$ are finite real numbers, the position coordinates of $P$ can be rounded to integer values in $DM$ for purpose of computational efficiency. At this stage \ie, without further processing, three cases may occur: (\textit{case1}) locations of $DM$ with just one point; (\textit{case2}) locations with more than one point from $P$; and (\textit{case3}) locations of $DM$ without corresponding point (``empty''). In this paper, which deals with the problem of obtaining depth-maps from a Velodyne HDL-64 sensor, the \textit{case1} occurs in approximately 6.7\% of pixel locations of the $DM$, in the \textit{case2} we have about 0.1\% of locations with more than a single point and finally, for the \textit{case3}, about 93.2\% of positions in $DM$ are empty (
unsampled). Those percentages were calculated from experimental data using 100 frames of the KITTI dataset (more details in Sect. \ref{sec:ROI_size}). Although negligible for one pixel-size (let's say a mask of $1 \times 1$), the \textit{case2} deserves particular attention as the number of sample points rapidly increases as the size of the area of interest, the local window, increases to $3 \times 3$, $5 \times 5$, and so on. This situation is better understood after viewing Fig. \ref{fig:foreground_back} where, due to the LIDAR's perspective viewing, range points pertaining to the cyclist, located in the foreground, and points belonging to the background are situated in the same $DM$'s location (the zoomed view in the bottom row of Fig. \ref{fig:1} illustrates this problem.). This yields large deviations for the average range value at a local window and is the main cause of errors in the depth map. As explained before, the set of measurement points from a LIDAR sensor, in the form of the point-cloud $PC$, 
are converted to discrete locations in a two-dimensional map $DM$, in pixel coordinates, and is represented by $P$. The main difference from other upsampling problems is that $DM$ has a very low density of points: of the order of 6.8\%. Therefore, the goal is to find a technique to estimate the value of $r$ in unsampled locations of $DM$ and keeping those estimated values of depth (range-distance to the LIDAR) consistent through the depth-map.

The discussion above allows us to concentrate on solving the \textit{case3} thus, we need a solution of estimating the desirable variable $r$ of the unsampled positions in $DM$ and, at the same time, the errors due to the boundaries regions (edges) should be minimized. Carrying out a solution of estimating depth values in locations without measurement points will result in a depth-map with higher resolution than the input; this is known as ``upsampling'' and can be formulated as a spatial estimation problem as detailed in the next sections.

\subsection{Defining the region of interest}
\label{sec:roi}

Spatial data interpolation, or estimation, is typically performed under the assumption that a local region of interest $R$ is previously defined and the desired point-value to be estimated is located within $R$; this is known as \textbf{local interpolation}. In a more general case, any polygon shape can be used to define $R$, however, in the problem considered here the most usual solutions are square regions (usually called mask or window). Nevertheless, for purpose of completeness, solutions based on Delaunay-triangles will also be considered. Delaunay-triangulation is effective in obtaining depth-maps with close to 100\% of density \ie, all locations with unknown depth-values are estimated, because this method interpolates all points regardless the distance between the points of a triangle. Except for Delaunay based methods, all the interpolation methods discussed in this paper are applied to estimate the range value of locations centred at the local and square window $R$ with size defined by $mr \times 
mr$ (in pixel units). The value of $mr$ is not simple to decide and has direct impact on the number of available points to estimate the variable of interest and consequently on the computational effort. Moreover, an important aspect to take into account is the minimum number of points in $R$ necessary to guarantee consistency, statistical significance and efficiency of the estimator. This paper will address this issue by experiments, investigating the spatial-resolution of the $DM$ (\ie, its density) and some statistics for increasing values of $mr$. The implementation consists in `moving' $R$ through the locations of $DM$ using the sliding window technique; then, all the points within $R$ are considered for estimating, locally, the depth value of the centre point of the window. A variety of interpolation (or estimation) methods can be applied to estimate the desired depth-value, some of them are described next.

\subsection{Local interpolation algorithms}
\label{sec:local_interpol}
The role of an interpolation method is to estimate values of range measurements, from a LIDAR, in both sampled and unsampled (empty) locations of the depth-map ($DM$). The locations in $DM$ are in pixel coordinates and the interpolation is performed locally \ie, restricted to a local region $R$. Let $\mathbf{x}_0 = (u,v)_0$ be the location of interest, which is the centre of $R$, and let $r^*_0$ be the variable to be estimated, that is, the range-distance $r$ at $x_0$. Given a finite set of measured points $\mathbf{p}_i=(\mathbf{x}_i,r_i), i=1,2,\cdots,n$, where $\mathbf{x}_i \in R \subset DM$, spatial interpolation can be formulated in terms of a function that weights the depth values $r_i$ of the points $\mathbf{p}_i$ according to spatial (position) based parameters $\Theta(\mathbf{x}_i)$ thus, $r^*_0 = f(r_i,\Theta(\mathbf{x}_i))$. There are many possibilities for $f(^.)$ to solve the problem but, for obvious 
reasons, we will address a limited number of methods. Besides, the following basic operators will be considered: sample average, minimum, maximum, median, nearest neighbor. The following three classes of interpolation techniques will be considered in this work: 1) inverse distance weighting ($IDW$), the simplest variant of the Shepard’s Method; 2) the Ordinary Kriging ($KRI$) and 3) a Polygon based method using Delaunay triangles ($DEL$).

The $IDW$ can be expressed as $r^*_0 = \sum_{i=1}^{n} \omega_{i}(\mathbf{x}) r_{i} /  \sum_{i=1}^{n} \omega_{i}(\mathbf{x})$, where $\omega_{i}(\mathbf{x}) = d_{i}^{-p}$, $d = ||\mathbf{x}_0 - \mathbf{x}_i||$ is a given distance function (a metric), and $p$ is a power parameter (positive real number). In a general form, the weights can be generalized to an arbitrary kernel function $K(^.)$ yielding, for Kernel-based methods, the expression: $r^*_0 = \sum_{i=1}^{n} K(\mathbf{x}_0,\mathbf{x}_i) r_{i}$. $KRI$ is an optimal linear estimator that estimates the value of a random function at the location of interest, $\mathbf{x}_0$, from samples $\mathbf{x}_i$ located in the local region of interest. The points $\mathbf{x}_i$ are weighted according to a covariance function or the equivalent semivariogram $\gamma(h)$. The parameters $\Theta$ in this method are the values of \textit{nugget}, \textit{sill} and \textit{range}; where $h$ represents the lag distance \ie, a distance measure between points. In the $DEL$ 
approach, the value of the unsampled point of interest, which lies within the triangle, can be estimated by different techniques. In this work, we use the Matlab classes \textit{delaunayTriangulation} and \textit{scatteredInterpolant} to interpolate the three points of a given triangle; the available techniques in \textit{scatteredInterpolant} are: linear, nearest neighbor, and natural neighbor interpolation.

In common with all those interpolation methods, the weighting is performed as function of the position of the sampled points $\mathbf{x}_i$, and the variable of interest ($r_i$) is not considered in the problem formulation. Conversely, the BF \cite{Tomasi} allows one of its weighting terms to be dependent of the variable of interest, in our case the range distance ($r_i$). For this reason, and due to the successful performance of Bilateral filtering in edge-preserving applications \cite{Bilateral}, we propose a modified version of BF to upsample depth maps.

\section{Edge-preserving on depth map upsampling}
\label{sec:Filtering}

In this section we briefly review the BF, a well-known edge-preserving filter, and propose a new range-weighting technique for upsampling depth-maps from 3D-LIDAR's data (as shown in Fig. \ref{fig:1}). For a detailed review on edge-preserving filters, in the domain of image processing, please see \cite{GuidedFilter}.

\subsection{Bilateral filter}
\label{sec:Bilateral}

Following the notations in Sect. \ref{sec:sect1}, Bilateral filtering \cite{Tomasi}\cite{Bilateral} can be expressed as follows:
\begin{equation}
\label{eq:bil1}
r^*_0 = \frac{1}{W} \sum_{\mathbf{x}_i \in R} G_{\sigma{_s}}(||\mathbf{x}_0 - \mathbf{x}_i||) G_{\sigma{_r}}(|r_{0} - r_{i}|) r_{i}
\end{equation}
where $G_{\sigma{_s}}$ weights the points $\mathbf{x}_i$ inversely to their distance to the position of interest $\mathbf{x}_0$, $G_{\sigma{_r}}$ controls the influence of the sampled points as function of their range values $r_{i}$, and finally $W$ is a normalization factor that ensures weights sum to one. In (\ref{eq:bil1}), we set $G_{\sigma{_s}}$ to be inversely proportional to the Euclidean distance between the center of the mask $R$ and the sampled locations $\mathbf{x}_i$, yielding
\begin{align*}
G_{\sigma{_s}} = \frac{1}{1+(||\mathbf{x}_0 - \mathbf{x}_i||)}.
\end{align*}
The influence of $G_{\sigma{_s}}$ is not very significant because the problem of jump discontinuities is caused by differences in range. On the other hand, $G_{\sigma{_r}}$ is the key component to be explored in order to provide improvement in the estimation of $r^*_0$, under the influence of discontinuities between foreground and background. A common form of weighting the range values, as in \cite{FusionIROS14}, is given by
\begin{align*}
G_{\sigma{_r}} = \frac{1}{1+(|r_0 -r_i|)}.
\end{align*}
However, as mentioned in Sect. \ref{sec:sect1}, the average percentage of centred pixel $\in R$ with range values is (only) 6.8\% therefore, the nearest value $r_0 = \min (r_i), \forall r_i \in R $ has been chosen at an unsampled location $\mathbf{x}_0$.

\subsection{Modified Bilateral filter}
\label{sec:modified_bilateral}

\begin{figure}[!t]
\centering
\includegraphics[width= 89.0 mm]{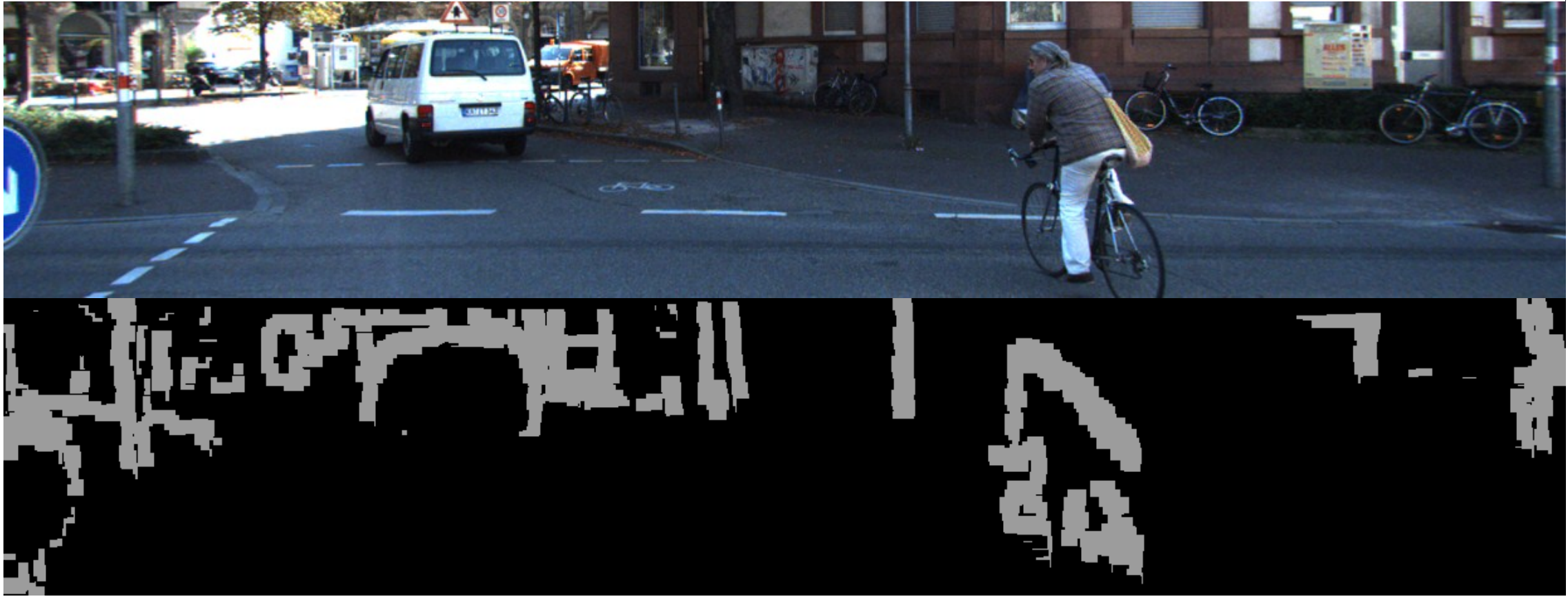}
\caption{The first row is an example-frame from the KITTI dataset and serves for visualization purpose; in the row below, the pixels in light-grey indicate the center of masks where $nc>1$ hence, corresponding to discontinuities/edges between foreground and background.}
\label{fig:edge_cluster}
\end{figure}

We propose a modification in the BF, henceforth indicated as $BF^*$, by expressing the weighting element $G_{\sigma{_r}}$ as a function of the `dispersion' of $r$ in the mask $R$. Assuming that an edge is characterized by a discontinuity in the range values, we propose to use a clustering algorithm to detect discontinuities and, if it is the case, the number of clusters ($nc$) will be at least two; therefore, an edge (or a discontinuity) occurs if $nc > 1$ (as shown in Fig. \ref{fig:edge_cluster}). The algorithm used to perform clustering is based on the popular DBSCAN \cite{dbscan}, which is a simple and effective algorithm that depends on two parameters, $\epsilon$ and $minPts$. The implementation of the DBSCAN should take into consideration six definitions, as detailed in \cite{dbscan}, and a distance function between points. In this work, we consider the distance function ($DF$) as given by:
\begin{align*}
DF = \vert \frac{r_{k}-r_{k+1}}{r_{k}+r_{k+1}}	\vert, \quad k=1,\cdots,nR, 
\end{align*}
\noindent
where $nR$ is the number of points $\in R$. If $DF>\epsilon$ \ie, a discontinuity has been detected, the occurrence of more than one cluster is likely true. A cluster ($s_{i}$) is accepted only if $minPts>1$. Ideally, the number of clusters corresponding to a window where an edge occurs should be $nc=2$ \ie, one clustered set of points belonging to the foreground and another to the background. We conducted experiments using the DBSCAN algorithm, with $\epsilon=0.08$ and $minPts=2$, and found out that $nc$ is equal to two in the majority of the regions where an edge occurs. Figure \ref{fig:edge_cluster} provides a visual display where the light-grey pixels correspond to regions with $nc>1$. In some circumstances, however, $nc$ is greater than 2 and it is not easy to detect the `optimal' boundary between foreground \textit{vs} background. Whenever $nc \geq 2$, the approach presented in this section considers at most two clusters; conversely, if $nc=1$ equ. \equ{bil1} is applied to all the points in $R$. When 
$nc > 1$ the $BF^*$ uses a ratio ($\lambda$) between the number of points of, at most, two clusters.

Let $np_1$ and $np_2$ be the number of points belonging to the clusters $s_1$ and $s_2$, where $\lambda = np_1/np_2$ is the ratio of interest. The variable $np_1$ corresponds to the cluster that has the closest average distance to the LIDAR (denoted cluster $s_1$); this holds regardless of the number of remaining clusters. However, if $nc>2$ and excluding the cluster $s_1$, then a decision process is carried out, to select $s_2$, according to the following rule: $s_2$ is chosen as the cluster with more points and, in case of clusters with the same number of points, then $s_2$ corresponds to the one with the closest average distance. Once the pair $(s_1,s_2)$ has been selected, then a threshold-based rule is applied to the ratio $\lambda \geq Thr$ in order to penalize $s_1$ or $s_2$. This rule is exclusive in the sense that one of the clusters will be excluded, therefore if $\lambda \geq Thr$, then the $BF^*$ will be run on the points belonging to $s_1$ else, only the points in $s_2$ will be considered. The 
key idea is to strongly penalize, based on the ratio $\lambda$, one of two clusters and, as consequence, only points belonging to one of the clusters will be considered in \equ{bil1}.

\section{Dataset and evaluation methodology}
\label{sec:eval}

Publicly datasets for purpose of LIDAR-based depth map evaluation and performance assessment are not available at the time of this writing. Therefore, to provide quantitative evaluation, we resort to the \textit{KITTI Stereo 2015} which provides groundtruth for disparity maps evaluation and benchmark. But, because that dataset was built to evaluate stereo systems, we had to find a solution that makes the evaluation of LIDAR-based depth maps possible; this is explained in the following section.

\subsection{KITTI dataset for depth-map evaluation}
\label{sec:kitti_eval}

\begin{figure}[!t]
\centering
\includegraphics[width= 89.0 mm]{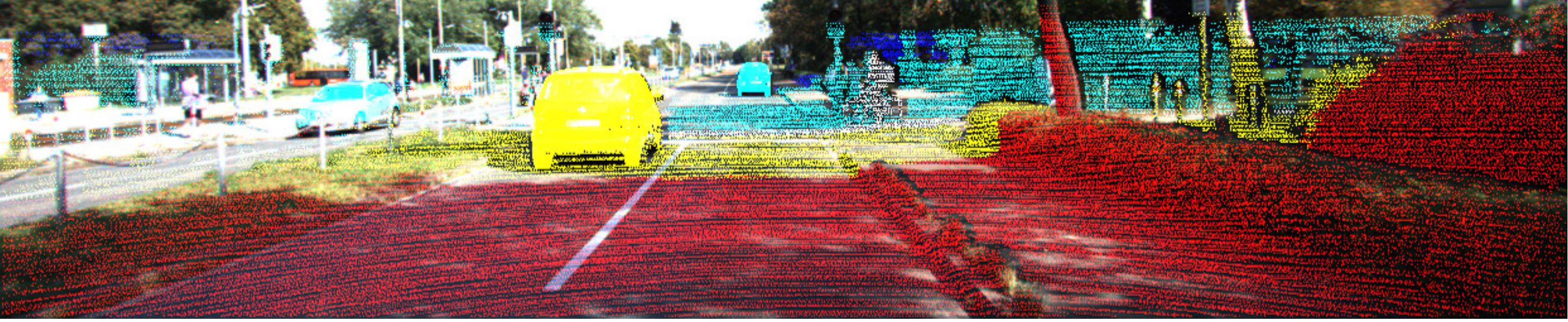}
\caption{An example of groundtruth point cloud, superimposed on a color image, with some objects (vehicles) having better resolution than the majority regions in the groundtruth. Those well delineated vehicles, that represent foreground objects, are a challenging situation for a solution using single LIDAR-based point cloud.}
\label{fig:gt_examples}
\end{figure}

The present \textit{KITTI Stereo 2015}\footnote{ \url{http://www.cvlibs.net/datasets/kitti/eval_scene_flow.php?benchmark=stereo} } comprises a total of 400 frames from stereo images, 200 for training with their corresponding disparity maps (the groundtruth), and 200 for benchmarking purposes, where each frame has two associated images: one from each camera of the stereo pair. The \textit{KITTI Stereo 2015} is part of the \textit{KITTI Vision Benchmark Suite}, being the latter composed of thousands of frames from different sensors: a Velodyne LIDAR, cameras, and GPS/IMU. The groundtruth for disparity evaluation was created using a process that incorporates a set of consecutive scans from the LIDAR (5 before and 5 after the actual frame of interest), where this sequence of point clouds were conveniently merged by a ICP technique as reported in \cite{Geiger2012CVPR}, followed by a manually correction step to rectify eventual ambiguities. Nevertheless, the actual \textit{KITTI Stereo 2015} dataset provides a 
more challenging and accurate groundtruth, described in \cite{Menze2015CVPR}, where ``objects'' (vehicles) in the scenes were recovered by fitting detailed CAD models to the point clouds. The consequence is a set of groundtruth frames where some objects are very well delineated as shown in Fig. \ref{fig:gt_examples}. To make possible the evaluation of depth-maps generated solely by LIDAR, a new dataset - using data from KITTI - had to be built. First, the groundtruth, originally in the form of disparity maps, has to be converted to depth maps by known geometry. To calculate depth $Y_{E}$ of a disparity map $Y_{I}$, it is required to know the values of the baseline $B$ and the focal length $f_{c}$, and the conversion is given by $Y_{E} = Bf_{c}/Y_{I}$. Once the \textit{KITTI Stereo 2015} made available image-frames and the corresponding disparity maps, it is necessary to find the LIDAR scans that match the groundtruth frames. In this work, we performed a non-exhaustive search in the ``Raw Data'' recordings of 
the \textit{KITTI Vision Benchmark Suite} and established the correspondence between 100 LIDAR scans and their counterpart in the groundtruth set in \textit{KITTI Stereo 2015}. Henceforth, the experiments presented in the next sections were carried out using this set of 100 scans.

\subsection{Methodology}
\label{sec:methodology}

The evaluation methodology adopted in this work is similar to the one used in the \textit{KITTI Stereo 2015} benchmark, excepting the following: the total number of frames used in the evaluation is 100, and the separation of training and testing set is not carried out once our approach does not depend on a learning strategy. In short, we adapted the C++ codes of the development kit package, from KITTI website, to run the evaluation routine on the dataset described above. Four performance measures, detailed in \cite{Menze2015CVPR}, are calculated: \textit{D-bg}, \textit{D-fg}, \textit{D-all}, and \textit{Density}. The meaning of these terms are:\\
- \textit{D-bg}: \% of outliers averaged over background regions;\\
- \textit{D-fg}: \% of outliers averaged over foreground (objects);\\
- \textit{D-all}: \% of outliers averaged over all groundtruth pixels;\\
- \textit{Density}: \% of the average number of pixels, in the depth-maps, with coincident positions of the groundtruth pixels.

Although the groundtruth, as already mentioned, has been obtained by a combination of 10 point clouds and by complementary object CAD-models, the number of unsampled pixels in the groundtruth depth-maps is still significant (as shown in Fig. \ref{fig:gt_examples}). So, the groundtruth depth maps are not 100\% dense \ie, the value of \textit{Density} is calculated considering only the sampled pixels \cite{Menze2015CVPR}.
\section{Experiments and evaluation}
\label{sec:Experiments}

This section describes the experiments conducted on the dataset detailed in Sect. \ref{sec:kitti_eval}, with the goals of assessing the performance of the local-region ($R$) and evaluating the approaches discussed in Sect. \ref{sec:local_interpol} and Sect. \ref{sec:Filtering}. The reported results are from the non-occluded set ($Disp\_noc\_0$). An appropriate solution to mitigate the errors caused by discontinuities between foreground and background will depend, primary, on the detection of the occurrence of such discontinuities. As mentioned in Sect. \ref{sec:Bilateral}, a BF-modified version, $BF^*$, was proposed to obtain consistent depth-map from solely LIDAR data.

\subsection{The role of the region of interest}
\label{sec:ROI_size}

The size of $R$, defined by $mr \times mr$ in pixel units as described in Sect.\ref{sec:roi}, controls the number of points (within $R$) to be used by a given local-interpolation approach. Table \ref{tab:roi} provides the results, averaged over the 100 frames of the dataset, for increasing values of $mr$. The average number of points in $R$ is denoted by $N_{ave}$, while $N_{max}$ indicates the maximum number of points. These values were obtained by applying a sliding-window strategy with step of 1 pixel. The density of the map, denoted by $D_{ens}$ and given in percentage, is calculated considering the region covered by the LIDAR's field-of-view (FOV). As shown in Fig. \ref{fig:1}, the LIDAR measurements cover a limited vertical FOV \ie, the upper part of the image plane is empty, and hence only the pixels below a certain `horizon line' (see Fig. \ref{fig:1}) are used to compute $D_{ens}$. A density of 100\% means that the window $R_{mr \times mr}$ had, in all locations of the sparse map, at least 1 point 
inside $R$. The `horizon line' was calculated by averaging, for each column of $DM$, the points of $P$ that have the smallest value in vertical-axis.

The KITTI Database provides the MATLAB/C++ utility package used in the evaluation of the algorithms. Density is evaluated against the KITTI groundtruth and, therefore, it follows a methodology which is not the same as the above. For that reason, the values of density (denoted by $Den^{*}$) as calculated by the KITTI evaluation package, shown in the last row of Table \ref{tab:roi}, are not the same of $D_{ens}$.

\begin{table}[t]
\caption{Statistics for increasing size of $R$. }
\label{tab:roi}
\centering
\scriptsize
\begin{tabular}{ m{3.8mm} | m{3.8mm} | m{3.8mm} |m{3.8mm} | m{3.8mm} | m{3.8mm} | m{3.8mm} | m{3.8mm} | m{3.8mm} | m{3.8mm} | m{3.8mm} }
\hline
$mr$		& 3	& 5 	& 7 	& 9 	& 11 	& 13 	& 15 	& 17 	& 19 	& 21 \\
$N_{max}$	& 6	& 11 	& 17 	& 23 	& 36 	& 42 	& 55 	& 64 	& 75 	& 89 \\
$N_{ave}$	& 0.61 	& 1.69 	& 3.31 	& 5.47 	& 8.17 	& 11.39 & 15.15 & 19.43 & 24.24 & 29.58 \\
$D_{ens}$	& 48.90 & 81.09	& 93.52 & 96.58	& 97.48 & 97.99 & 98.33 & 98.59 & 98.80 & 98.98 \\
$Den^{*}$	& 58.17 & 87.17	& 96.66 & 98.74	& 99.26 & 99.52 & 99.67 & 99.77 & 99.82 & 99.86 \\
\hline
\end{tabular}
\end{table}

\subsection{Evaluation of local-spatial interpolation algorithms}
\label{sec:exp_methods}

The first experiments conducted in this work involve methods that do not depend on the size of $R$, which are the cases of the Nearest ($NEA_{nei}$) operator and the Delaunay-based techniques using linear ($DEL_{lin}$), nearest neighbor ($DEL_{nei}$), and natural neighbor ($DEL_{nat}$) interpolation. In terms of error performance, as shown in Table \ref{tab:res_noROI}, $NEA_{nei}$ and $DEL_{nea}$ achieved comparable results, however the main difference resides in the fact that the density resulted from $NEA_{nei}$ varies with the window size ($mr$), while the Delaunay-based approaches attain the same value of density, being equal to 99.96\%.

\begin{table}[!th]
\caption{Evaluation results using methods where the errors are independent of $R$ size.}
\label{tab:res_noROI}
\centering
\normalsize
\begin{tabular}{l | c | c | c }
\hline
{\bf Method} 	& \textit{D1-fg} & \textit{D1-bg} & \textit{D1-all} \\
\hline
$NEA_{nei}$		& 17.47 \% & 3.78 \% & 5.53 \%  	\\
$DEL_{lin}$ 		& 22.05 \% & 4.15 \% & 6.48 \%  	\\
$DEL_{nea}$ 		& 17.10 \% & 3.82 \% & 5.55 \%  	\\
$DEL_{nat}$ 		& 23.54 \% & 4.21 \% & 6.73 \%  	\\
\end{tabular}
\end{table}

\begin{table}[!th]
\caption{Evaluation results for $mr = 13$.}
\label{tab:res_final}
\centering
\normalsize
\begin{tabular}{l | c | c | c }
\hline
{\bf Method} 	& \textit{D1-fg} & \textit{D1-bg} & \textit{D1-all} \\
\hline
$BF^*$		& 8.23 \%  & 2.63 \% & 3.35 \%  		\\
$MIN$ 		& 7.57 \%  & 4.20 \% & 4.63 \%  		\\
$BF$ 		& 14.64 \% & 3.32 \% & 4.77 \%  	\\
$MED$ 		& 20.37 \% & 4.91 \% & 6.88 \%  	\\
$IDW$ 		& 25.84 \% & 4.41 \% & 7.14 \%  	\\
$KRI$ 		& 25.77 \% & 4.54 \% & 7.25 \%  	\\
$AVE$ 		& 26.67 \% & 4.77 \% & 7.56 \%  	\\
$MAX$ 		& 34.12 \% & 15.37 \% & 17.76 \%  	\\
\end{tabular}
\end{table}

\begin{figure}[!t]
\centering
\subfigure[D1-fg]
{\includegraphics[width=81.2mm]{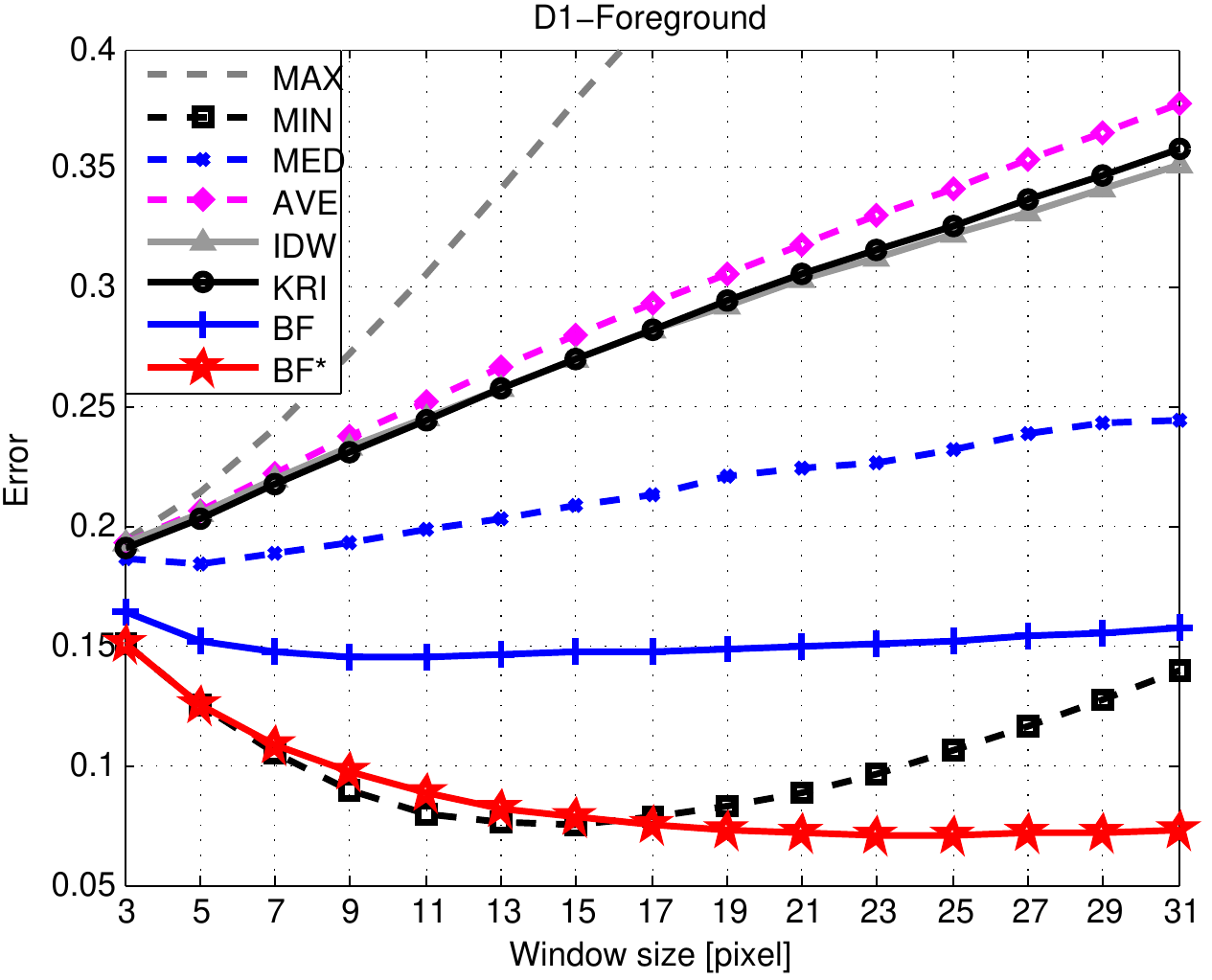}
\label{fig:d1_foreground}}
\hfil
\subfigure[D1-bg]
{\includegraphics[width=81.2mm]{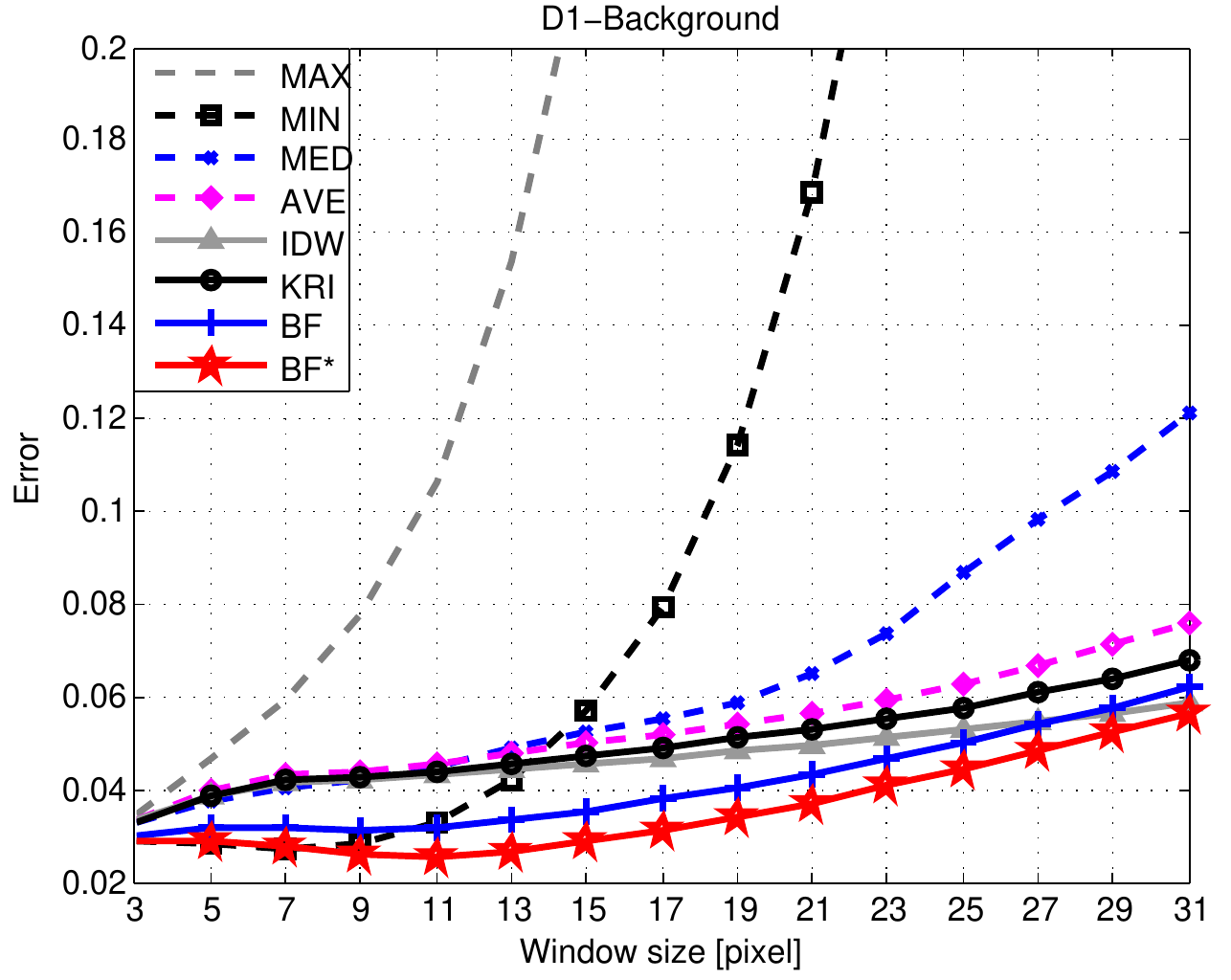}
\label{fig:d1_background}}
\hfil
\subfigure[D1-all]
{\includegraphics[width=81.2mm]{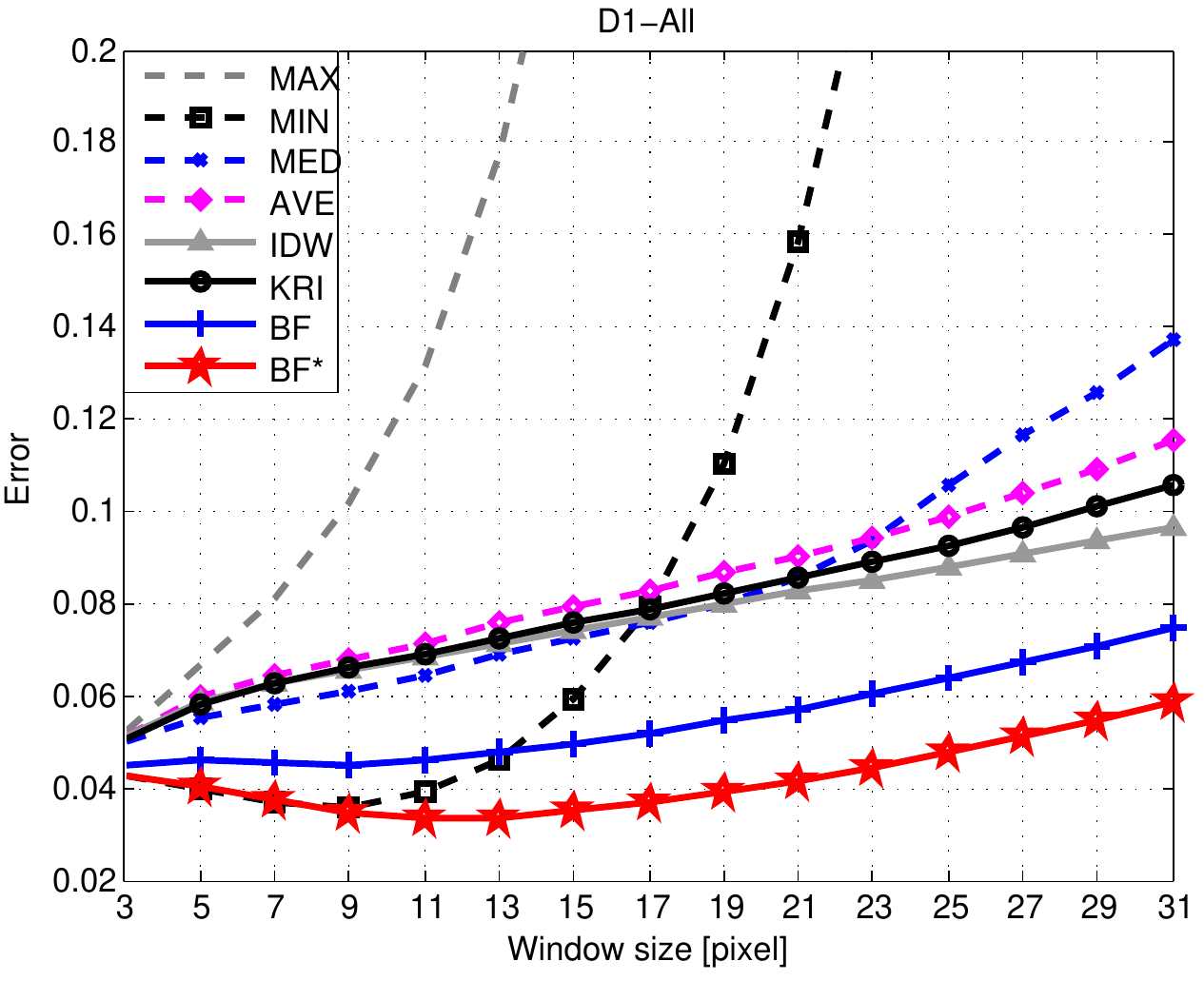}
\label{fig:d1_all}}
\caption{Error curves, as function of the window size $mr$, for the methods and techniques addressed in this work. The curves for the Delaunay-based techniques and the nearest-neighbor are omitted because $mr$ does not influence their error performances.}
\label{fig:results_error}
\end{figure}

The methods $IDW$ and $KRI$, the operators sample average ($AVE$), minimum ($MIN$), maximum ($MAX$) and median ($MED$), and also the BF and $BF^*$ have different performances for different values of $mr$. For this reason, and to show the relationship between the errors and $mr$, Figure \ref{fig:results_error} provides the values of the error measures (\textit{D1-bg}, \textit{D1-fg} and \textit{D1-all}) for increasing values of window size.
\subsection{Discussion}
\label{sec:discussion}

Considering the results shown in Fig. \ref{fig:results_error} the error (\textit{D1-fg}) on the foreground objects (Fig. \ref{fig:d1_foreground}) demonstrated to be the most challenging case, particularly for the operators $MAX$, $AVE$, and methods $KRI$, $IDW$, where the error increases monotonically with $mr$. For the median operator ($MED$), the situation is also not favorable. The $MIN$ shows a good behavior up to $mr = 15$, while $BF$ and $BF^*$ are relatively robust for all values of $mr$, although the proposed $BF^*$ attained the best results. In terms of background-errors (\textit{D1-bg}), depicted in Fig. \ref{fig:d1_background}, most of the methods are generally satisfactory except, clearly, the operators $MAX$, $MIN$ and $MED$. Finally, and as consequence of the combined errors (\textit{D1-all = D1-fg + D1-bg}), from Fig. \ref{fig:d1_all} it is possible to conclude that $BF^*$ achieved the lowest error among the implemented methods. Based on the values of density, provided in Table \ref{tab:roi}, 
and considering the error plots in Fig. \ref{fig:results_error}, the `optimum' value of the window size is $mr=13$. Therefore, Table \ref{tab:res_final} shows the values of error, in percentage and for a window size of $13 \times 13$, which facilitates the comparison of results among the methods and techniques, including those reported in Table \ref{tab:res_noROI}. 

\begin{figure*}[!ht]
\centering
\includegraphics[width= 184.0 mm]{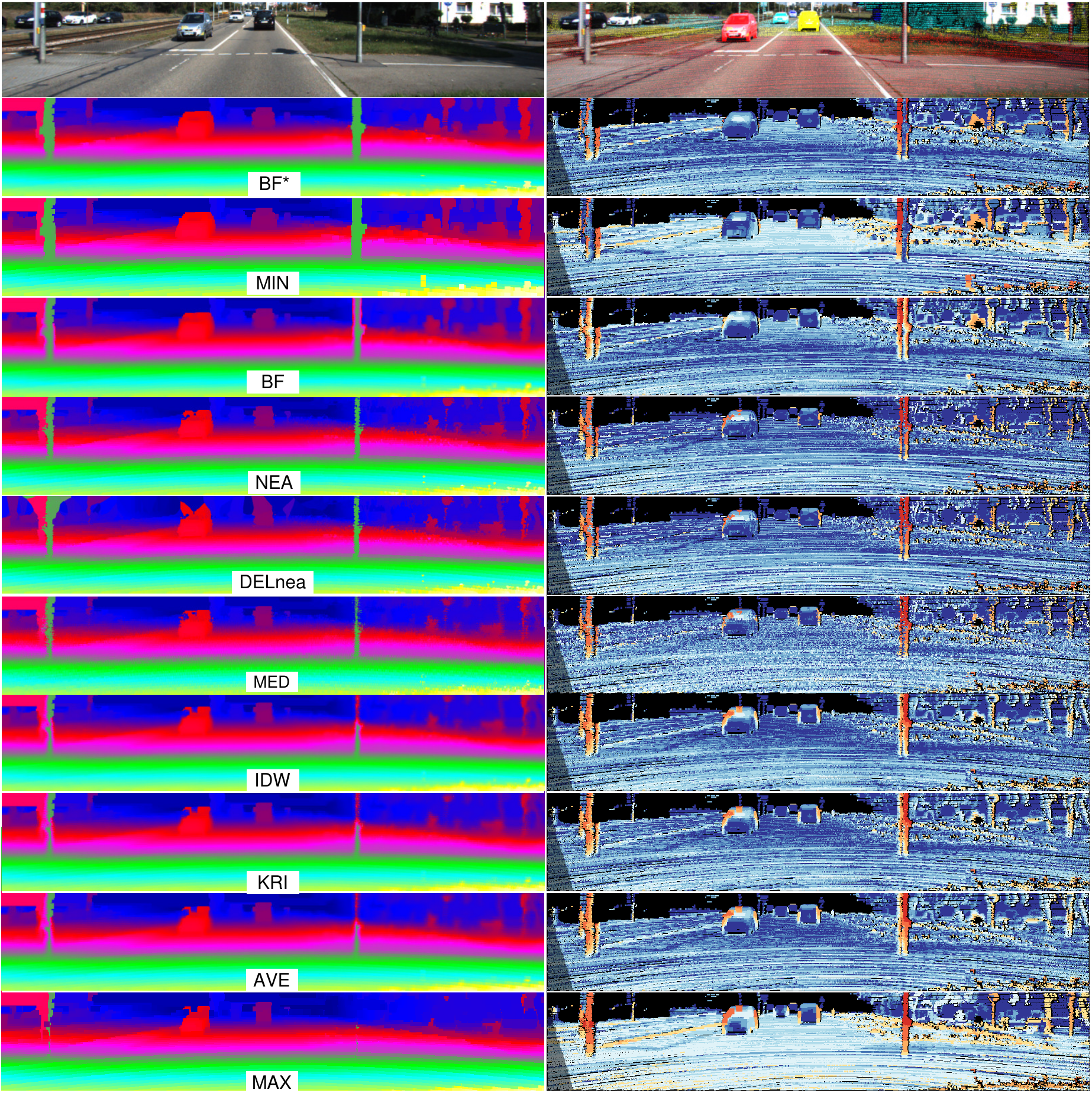}
\caption{Qualitative results from the techniques and methods discussed in this paper, for a mask of size $13 \times 13$ pixels. In the top-left part a color image of the scene is provided, while the top-right shows the groundtruth (point cloud) superimposed in the image. The depth-maps \ie, the output of the algorithm implementations, and the error maps are shown in left and right columns respectively.}
\label{fig:qualitative_res}
\end{figure*}

In terms of qualitative results, Fig. \ref{fig:qualitative_res} shows, for a given frame, the resulting depth-maps (left part) and the error images (in the right). The color image and the groundtruth are provided to facilitate the analysis; the output depth-maps are shown using a colormap which is proportional to the distance, while the error images were mapped using the KITTI dev-kit \cite{Menze2015CVPR}. Notice that the top part of the depth-maps where removed because the LIDAR's FOV does not cover the entire frame (as discussed in Sect. \ref{sec:ROI_size}).

\section{Conclusion}
\label{sec:conclusions}

A high-resolution, LIDAR-based only, depth mapping approach is presented in this work as a promising solution to be used in sensory perception systems, as part of applications such as: road detection, object recognition and tracking, environment modeling, cooperative perception. The approach is based on the Bilateral Filter (BF) and contributes with a technique that influences the weighting range-term of the BF. Experiments using the KITTI database were carried out to assess the performance of the proposed approach as well as other usual interpolation techniques, namely: Kriging, IDW, Delaunay interpolation, Median, Nearest neighbor, and others. From the experimental results reported in this paper, the proposed approach, denoted by $BF^*$, attained the best results among the methods and techniques tested.


%
%
\section*{Acknowledgment}
\footnotesize{This work has been supported by FCT and COMPETE under projects ``AMSHMI2012-RECI/EEIAUT/0181/2012'' and ``UID/EEA/00048/2013''.}

\bibliographystyle{ieee}
\bibliography{Refs.bib}

\end{document}